\documentclass[10pt,twocolumn,letterpaper]{article}

\usepackage{iccv}
\usepackage{times}
\usepackage{epsfig}
\usepackage{graphicx}
\usepackage{amsmath}
\usepackage{amssymb}

\usepackage{algorithm}
\usepackage{algorithmic}
\usepackage{caption}
\usepackage{subcaption}
\usepackage{pgfplots}
\usepackage{comment}
\usepackage{tabulary,multirow}

\newcolumntype{x}[1]{>{\centering\arraybackslash}p{#1pt}}

\newlength\savewidth\newcommand\shline{\noalign{\global\savewidth\arrayrulewidth
  \global\arrayrulewidth 1pt}\hline\noalign{\global\arrayrulewidth\savewidth}}

\makeatletter\renewcommand\paragraph{\@startsection{paragraph}{4}{\z@}
  {.5em \@plus1ex \@minus.2ex}{-.5em}{\normalfont\normalsize\bfseries}}\makeatother

\usepackage{array}
\newcolumntype{L}[1]{>{\raggedright\let\newline\\\arraybackslash\hspace{0pt}}m{#1}}
\newcolumntype{C}[1]{>{\centering\let\newline\\\arraybackslash\hspace{0pt}}m{#1}}
\newcolumntype{R}[1]{>{\raggedleft\let\newline\\\arraybackslash\hspace{0pt}}m{#1}}

\usepackage[pagebackref=true,breaklinks=true,colorlinks,bookmarks=false]{hyperref}

\iccvfinalcopy % *** Uncomment this line for the final submission

\def\iccvPaperID{4169} % *** Enter the ICCV Paper ID here

\ificcvfinal\fi
\begin{document}

%%%%%%%%%%%%%%%%%%%%%%
% TITLE
%%%%%%%%%%%%%%%%%%%%%%
\title{Adversarial Learning of General Transformations for Data Augmentation}

%%%%%%%%%%%%%%%%%%%%%%
% AUTHORS
%%%%%%%%%%%%%%%%%%%%%%
\author{Saypraseuth Mounsaveng\\
\'{E}cole de Technologie Sup\'{e}rieure\\
Montr\'{e}al, Canada\\
{\tt\small saypraseuth.mounsaveng.1@etsmtl.net}
\and
David Vazquez\\
Element AI\\
Montr\'{e}al, Canada\\
{\tt\small dvazquez@elementai.com}
\and
Ismail Ben Ayed\\
\'{E}cole de Technologie Sup\'{e}rieure\\
Montr\'{e}al, Canada\\
{\tt\small ismail.benayed@etsmtl.ca}
\and
Marco Pedersoli\\
\'{E}cole de Technologie Sup\'{e}rieure\\
Montr\'{e}al, Canada\\
{\tt\small marco.pedersoli@etsmtl.ca}
}

\maketitle
%\thispagestyle{empty}

%%%%%%%%%%%%%%%%%%%%%%
% ABSTRACT
%%%%%%%%%%%%%%%%%%%%%%
\begin{abstract}
Data augmentation (DA) is fundamental to prevent large convolutional neural networks from overfitting, especially with a limited amount of training samples. In images, DA is usually based on heuristic transformations, such as image flip, crop, rotation or color transformations. %, known to be dataset dependent.
Instead of using predefined transformations, DA can be learned directly from the data. Existing methods either learn how to combine a set of predefined transformations or train a generative model used for DA. Our work combines the advantages of the two approaches. It learns to transform images with a spatial transformer network combined with an encoder-decoder architecture in a single end-to-end fully differentiable network architecture. Both parts are trained in an adversarial way so that the transformed images still belong to the same class, but are new, more complex samples for the classifier. Our experiments show that, when training an image classifier, our approach is better than previous generative data augmentation methods, and comparable to methods using predefined transformations, which require prior knowledge.

%Existing methods either learn how to combine a set of predefined transformations or learn a generative model used for data augmentation. The first approach requires prior information and can be slow to train if chosen transformations are not differentiable, which precludes the use of backprop. 
% The second approach can in principle learn any transformation, but is prone to overfitting on reduced training sets.\\
% Our work combines the advantages of the two approaches and brings following improvements: i) Instead of predefined transformations, we learn useful affine transformations with a spatial transformer network, which is fully differentiable; ii) Instead of generating an image from scratch, as in the generative approaches, we learn to transform it with an encoder-decoder architecture, which is easier but still generic to any transformation.
% Both parts are trained in an adversarial way such that the transformed image still belongs to the same category, but is a new, more complex sample for the classifier. 
% Comprehensive experiments show that our approach is better than previous generative data augmentation methods, and comparable to predefined transformation methods, while enjoying a faster fully differentiable end-to-end training.

\end{abstract}

%%%%%%%%%%%%%%%%%%%%%%
% INTRO
%%%%%%%%%%%%%%%%%%%%%%
\section{Introduction}
\label{sec:intro}
% Deep learning need much data. Data augmentation reduce overfitting
Convolutional neural networks have shown impressive results in visual recognition tasks~\cite{hu2018squeeze}. However, for a proper training and good performance, they require large labeled datasets~\cite{DBLP:conf/eccv/MahajanGRHPLBM18, sun2017revisiting}. If the amount of training data is small, regularization techniques~\cite{srivastava2014dropout, krogh1992simple} can help the model avoid overfitting. Among these techniques, data augmentation seems to be the most effective in improving the final performance of the network~\cite{hernandez2018data, perez2017effectiveness}.
\begin{figure}[htb]
\centering
  \begin{subfigure}[b]{.46\columnwidth}
    \centering
    \includegraphics[width=\textwidth]{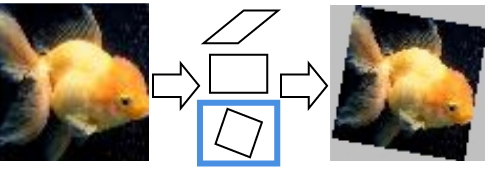}
    \caption{Predefined transformations}\label{fig:preselection}
\end{subfigure}
\hspace{0.1cm}
  \begin{subfigure}[b]{.43\columnwidth}
    \centering
    \includegraphics[width=0.97\textwidth]{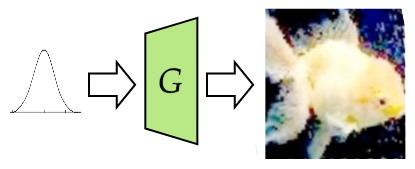}
    \caption{GAN image generation}\label{fig:gan_generation}
  \end{subfigure}\\%
  \begin{subfigure}[b]{\columnwidth}
    \centering
    \includegraphics[width=0.95\textwidth]{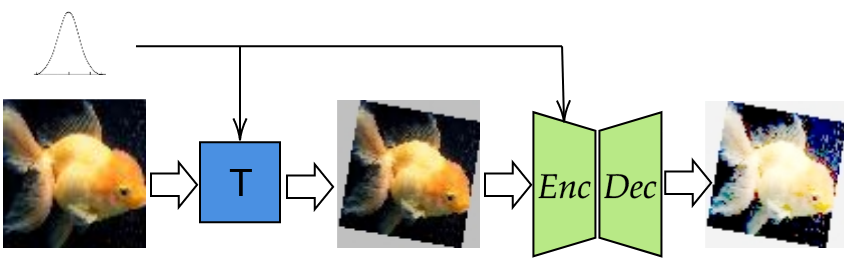}
    \caption{Our approach}\label{fig:our_approach}
  \end{subfigure}%
  \caption{\textbf{Automatic data augmentation approaches.} 
  \\(a) Sequence of predefined transformations are automatically selected. (b) GAN model generates new images from the same distribution.  (c) Our approach combines the advantages of the these two methods by combining affine transformations generated by a spatial transformer network and a transformation generated by a convolutional encoder decoder model. %Given an input image, the generator first performs a global transformation using a spatial transformer network followed by more localized transformations using a convolutional encoder-decoder network.
  }\label{fig:overview}
\end{figure}

% Problems of data augmentation
In images, data augmentation consists in applying predefined transformations such as geometric transformations like flip or rotations and color changes ~\cite{krizhevsky2012imagenet, Ciresan:2012g}. This approach works quite well and provides a consistent improvement of the accuracy when training a classifier. However, selecting the right transformations require prior knowledge and chosen transformations are dataset dependent. For instance, if flipping an image horizontally makes sense for natural images, it produces ambiguities (\eg 2 and 5) on number datasets.

% Current data augmentation methods
% Several recent studies have investigated automatic data augmentation learning to avoid the manual selection of transformations. This can be achieved in different ways. The first one is to define a large enough set of transformations and learn how to combine them~\cite{ratner2017learning}. This approach works quite well, but relies on predefined transformations, which prevents the model from learning other transformations that could further improve the performance of a classifier. Additionally, when transformations are discrete, they need to be learned and combined with a reinforcement learning approach generally expensive to train. Alternatively, another approach for automatic data augmentations is to generate new samples from the probability distribution of the data, $p(X)$, e.g., via a GAN model~\cite{tran2017bayesian}. When the training data is reduced, however, it becomes very difficult to train a good generative model. Therefore, methods based on predefined transformations typically outperform the generative ones.  
Several recent studies have investigated automatic data augmentation learning, to avoid the manual selection of  transformations. \cite{ratner2017learning} defines a large set of transformations and learns how to combine them (Fig.\ref{fig:overview}(a)). This approach shows good results, but is based on predefined transformations, which prevents the learning of other transformations that could be useful for the classifier. Alternatively , \cite{chongxuan2017triple} and \cite{tran2017bayesian} generate new samples via a generative adversarial networks (GAN) based model from the probability distribution of the data $p(X)$ (Fig.\ref{fig:overview}(b)). % and ~\cite{antoniou2018augmenting} learn transformations of images, instead of generating images from scratch. %, but not fined tuned to any final task.
Those methods show their limit when the number of training samples is low, as it is difficult to train a good generative model with a reduced training dataset.
\cite{hauberg2016dreaming} learn the natural transformations existing in a dataset by aligning pairs of samples from the same class. This approach is efficient on easy datasets like MNIST but seems not applicable on more complex datasets.

Our work combines the advantages of generative models and transformations learning approaches in a single end-to-end network architecture.  %(Fig.\ref{fig:overview}(c)). % that we call Data Augmentation Network (DAN).
First of all, instead of learning to generate samples, our model learns to generate transformations of a given sample. In other words, instead of generating samples from $p(X)$, we learn to generate samples from $p(\hat{X}|X)$, with $X$ a training data point, which is easier, especially when the training data is reduced.
As shown in Fig.~\ref{fig:overview}(c), we propose an approach that combines a first transformation defined by an affine matrix with a transformation defined by a convolutional encoder-decoder architecture. In practice, we find that the affine transformation learns global image transformations, while the encoder-decoder architecture learns more localized transformations regarding local image distortions and color changes. Thus, the combination of the two leads to a general distribution of transformation that can be applied to any image-based training data and is not specific to a given domain or application.

%The predefined global 
Secondly, affine transformations are learned by an adaptation of spatial transformer network STN~\cite{jaderberg2015spatial}, so that the entire architecture is differentiable and can be learned with standard back-propagation. In its original use, the purpose of STN is to learn to transform the input data so that it becomes invariant to certain transformations. In contrast, our approach uses STN to generate a distribution of augmented samples in an adversarial way. We experimentally show that our approach is more effective in improving the classifier accuracy.
Finally, we show that for optimal performance, it is important to jointly train the generator of the augmented samples with the classifier in an end-to-end fashion. By doing that, we can also add an adversarial loss between the generator and classifier such that the generated samples are difficult, or adversarial, for the classifier. This further increases the final classifier accuracy.

We test our approach on MNIST, fashion MNIST, SVHN and CIFAR-10 datasets, both in full dataset and low-data regime. Our empirical results show that (i) each component of the network is important for optimal performance; and that (ii) for a given classifier architecture, our method outperforms the hand-defined data augmentation and most of the previous methods.

To summarize, the contributions of this paper are:
i) We propose a data augmentation network that is fully differentiable, trainable end-to-end, and can significantly improve the performance of any image-based classifier; 
ii) We devise STN in an adversarial way that together with an encoder-decoder architecture is able to learn a distribution of general transformations for augmenting the training data; 
iii) We experimentally show that, for data augmentation, learning image transformations is better than generating images from scratch and that learning data augmentation and classification jointly is more effective than in two separate tasks.

%%%%%%%%%%%%%%%%%%%%%%
% RELATED WORK
%%%%%%%%%%%%%%%%%%%%%%
\section{Related Work}
\label{sec:related_work}
\paragraph{Standard Data Augmentation.}
Data augmentation is an efficient regularizer for improving the performance of visual recognition methods~\cite{hernandez2018deep}, especially when dealing with small training sets prone to overfitting~\cite{Wagner2013LearningCN}.
Data augmentation is based on specific domain knowledge about data transformations useful for an end task while keeping the semantic meaning of the data. For natural image classification, the standard form of data augmentation is affine transformations like flip, rotation or color changes~\cite{perez2017effectiveness}. However, more complex transformations such as occluding parts of an image~\cite{devries2017improved} or blending two or more images~\cite{zhang2018mixup, lemley2017smart} were also proposed. 
Finally, \cite{devries2017data} uses a form of data augmentation by adding noise, interpolating, or extrapolating between samples in the feature space, instead of in the input space.
Adding the transformed samples in the training data can highly improve the end task performance, but, especially with new tasks and datasets, it is not clear which transformations are helpful or, on the opposite, harmful.

\paragraph{Model-based Transformations.}
Recently, some approaches tried to make the data augmentation automatic, to avoid the manual selection of transformations.
%by learning either a composition of predefined transformations, or by using a domain specific model to generate new samples.
Ratner~\etal~\cite{ratner2017learning} propose to learn a sequence of predefined transformations to generate new samples. By using a generative adversarial network (GAN), the generated samples are enforced to be close to the training data distribution. This approach works well, but is limited to a set of predefined transformations, which prevents other possibly helpful transformations to be learned. Additionally, as transformations might not be differentiable, optimization is performed with a reinforcement learning approach. This can make the training more difficult and slow. %MP: we do not say anything about training speed...

Sixt~\etal~\cite{sixt2018rendergan} and Shrivastava~\etal~\cite{shri2017learning} generate augmented samples from initial 3D models of the data and refine them with GANs. These approaches work quite well, but need a strong prior knowledge of how to generate the initial 3D view. We follow a similar approach of first generating a global transformation and then refining it, but without using any domain specific 3D model. Instead, we learn a transformation from the given samples with a spatial transformer network~\cite{jaderberg2015spatial}. This is more difficult because the transformations are learned, but more generic and applicable to any dataset.

Hauberg~\etal~\cite{hauberg2016dreaming} learn class specific transformations by considering pairs of samples within a class and learning the distribution of the transformations morphing one element of the pair to the other. This approach seems applicable only to simple datasets like MNIST.\\
In the context of medical imaging, Zhao~\etal~\cite{zhaocvpr2019} learn distributions of spatial and appearance transformations by aligning labeled and unlabeled samples to create new synthetized labeled samples. Those new labeled samples are used to improve the performance of an image segmentation models for brain MRIs.
Finally, Peng~\etal~\cite{peng2018jointly} propose to train jointly a data augmenter (again based on pre-defined transformations) and an end-task network for human pose estimation. 
%By training the augmenter in an adversarial way, it enforces the creation of difficult samples that will contribute to a better performance of the final model.

\paragraph{Adversarial Training.}
Goodfellow~\etal~\cite{43405} showed how to induce a trained neural network to perform a wrong classification by minimally changing the image. They compute the gradient of the loss of the network with respect to the image pixels and use it to modify only the most influential pixels. This results in new images (adversarial) that are almost indistinguishable for humans, but wrongly classified by the net. Miyato \etal ~\cite{miyato2018virtual} extended the approach to unlabelled samples. The adversarial images can be added to the training data to improve the robustness to adversarial examples. This also improves the classifier accuracy and can be seen as a form of data augmentation.
However, this method does not fully exploit the data augmentation power as the learned transformations are constrained to small changes in order to maintain unaltered the appearance of the image.

\paragraph{Generic Transformations with GANs.}
Multiple approaches use generative models to generate the augmented samples at pixel level with a convolutional encoder-decoder architecture. In theory, this is more powerful and flexible than defining a set of predefined transformations, provided that the dataset is large enough to learn the generative model.
For instance, Mirza~\etal~\cite{mirza2014conditional} and Odena~\etal~\cite{DBLP:conf/icml/OdenaOS17} proposed to generate images conditioned on their class, which could be directly used to augment a dataset. 
CatGAN~\cite{springenberg2016iclr} on the other hand, performs unsupervised and semi-supervised learning as a regularized information maximization problem~\cite{krause2010discriminative} with a regularization based on the generated samples. 
Also based on GAN, but directly used for data augmentation is the model proposed by Antoniou~\etal~\cite{antoniou2018augmenting}. In this case the authors condition directly on a given image. From our experiments this method seems to produce sub-optimal results because the generation is performed independently from the classification.
\cite{salimans2016improved} trains a discriminator coupled to a classifier by adding an additional class to the classifier for generated images. In this case, the unlabeled data generated by the generator can be seen as a form of data augmentation. \cite{zhang2018dada} extends this idea to low data regime by using a finer grain for the classifier. Instead of using K+1 classes as in \cite{salimans2016improved}, the classifier in this model uses in 2K classes, K classes for real data and K classes for generated data. 
% Triangle GAN~\cite{gan2017triangle}, 

Triple GAN~\cite{chongxuan2017triple} and Bayesian Data Augmentation~\cite{tran2017bayesian} train a classifier jointly with the generator. We follow the same strategy. 
However, these models are based on the direct generation of samples from noise, and, as we show in our experiments, this seems more difficult than transforming a given image (as we do), especially when the training data is reduced.
Notice that most of the presented GAN models are designed for semi-supervised learning. Instead, our aim is to train with a reduced dataset, without additional (non annotated) images. This is a more challenging task and therefore a direct comparison is difficult and out of the scope of this work.

%%%%%%%%%%%%%%%%%%%%%%
% MODEL
%%%%%%%%%%%%%%%%%%%%%%

\section{Our model}
\label{sec:model}

%\paragraph{Motivation.}
\label{subsec:motivation}
% In this section, we present the architecture of our proposed model and explain the motivation behind each element.
In this work, we aim to improve the performance of an image classifier by augmenting the training dataset with samples synthetized by transforming the initial dataset with learned transformations.
%We focus on the case of fully labeled datasets and leave the additional use of unlabeled data (semi-supervised learning) out of scope. %MP:already mentioned just at the end of previous paragraph

%Let $\{(x_i, y_i)\}$ be a set of input images and their class labels. 
Our goal is to learn a distribution of image transformations $\mathcal{T}$, so that given an input image $x_i$, $\mathcal{T}(x_i)$ represents all image transformations such that the semantic meaning of the image, \ie its class $y_i$, is preserved. We expect this distribution to be the optimal set of transformations in order to augment the training data of a given classifier $C$. 
%$\mathcal{T}(x_i)$ has the same label $y_i$ as $x_i$ and the inclusion of $\mathcal{G}(x_i)$ in the training of a classifier $C$ increases its accuracy. To enforce the learning of a distribution of transformations, we define a random noise vector $z$ drawn from a Gaussian distribution as support for our transformations, which changes the formulation of the transformations to $\mathcal{G}(x_i, z)$.
%To learn a distribution $\mathcal{G}(x_i, z)$, that is useful for $C$, we introduce the additional constraint that $\mathcal{G}(x_i, z)$ should not contain the identity transformation $\mathcal{I}$.

\begin{figure}[!tb]
\centering
\begin{subfigure}[b]{.62\columnwidth}
\includegraphics[width=\textwidth]{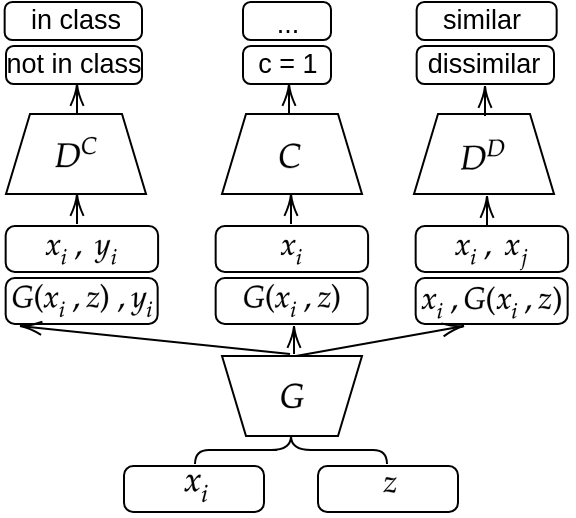}
\caption{Model architecture.}
%\label{fig:model}
\end{subfigure}
\begin{subfigure}[b]{.35\columnwidth}
\centering
\includegraphics[width=\textwidth]{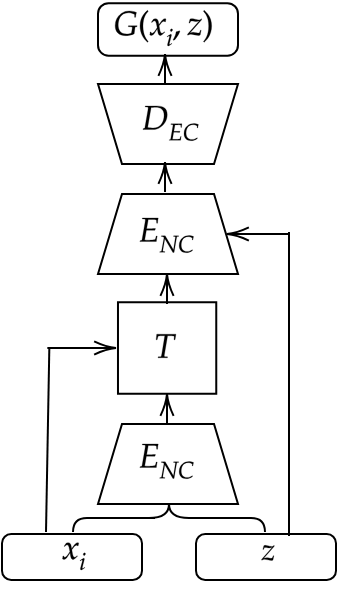}
\caption{Architecture of G.}
%\label{fig:generator}
\end{subfigure}
\caption{\textbf{Our model.} (a) A classifier $C$ receives augmented images from a generator $G$ challenged by two discriminators $D^C$ and $D^D$. The class discriminator $D^C$ ensures that the generated image $G(x_i, z)$ belongs to the same class as the input image $x_i$ with label $y_i$. The dissimilarity discriminator $D^D$ ensures that the transformed sample $G(x_i, z)$ is dissimilar from the input sample $x_i$ but similar to a sample $x_j$ from the same class. (b)  Given an input image $x_i$ and a noise vector $z$, our generator first performs a global transformation using a spatial transformer network followed by more localized transformations using a convolutional encoder-decoder network.}
\label{fig:models}
\end{figure}

To learn this distribution, we propose the GAN based architecture shown in Fig.~\ref{fig:models}(a). This architecture involves four modules: a generator $G$ to transform an input image, two discriminators $D^C$ and $D^D$ to impose constraints on the generated samples and a classifier $C$ that will perform the final classification task. 
% In contrast to most of the GAN models, instead of generating an input image, we learn a transformation of the image. This seems to be easier than learning to generate an image from scratch and leads to better classification results.
% The generator is supported by two discriminators. The first one is the class discriminator $D^C$, which ensures that the generated sample stays in the same class as the input sample. The second one is the dissimilarity discriminator $D^D$, which ensures that the transformed sample is different from the input sample. This is necessary to prevent the generator from learning the identity transformation, which would not help the classifier.
% Training our model consists in finding the equilibrium in a multiple two-player game. Indeed we solve jointly the adversarial game between $G$ and $D^C$, $G$ and $D^D$ and finally between $G$ and $C$. In contrast to previous GAN methods that use a classifier (\eg Triple GAN~\cite{chongxuan2017triple} and Bayesian DA~\cite{tran2017bayesian}), we introduce an additional loss pushing the generator to produce images that are difficult to classify, and this also helps to improve the classifier. 
In the following paragraphs we formally describe in detail each part of the model. %, while in Sec.~\ref{sub:ablation} we show their individual contribution with an ablation study.

 % Generator
\paragraph{Generation of augmented samples.}
The role of the generator $G$, is to learn the distribution $\mathcal{T}$ of the transformations of the input images that are the most useful to train the classifier $C$. In our intuition, learning an image transformation instead of learning a mapping from noise to image to generate new samples (as previous work), is an easier task in low data regime.\\
As shown in Fig.~\ref{fig:models}(b) the generator is composed of two elements: a Spatial Transformer Network (STN) %(in the figure represented as the combination of an encoder $E_{NC}$ and T as a combination of the STN grid generator and sampler) 
that learns global affine transformations and a U-Net (\cite{ronneberger2015u}) variant that can learn, in principle, any other transformation. 
While in the original paper the spatial transformer module was used for removing invariances from the input data, the proposed model generates transformed samples (controlled by $z$) in an adversarial way.

The entire transformations of an input image $x_i$ and a random noise vector $z$ can be formulated as:
\begin{equation}
\label{eq:gen}
G(x_i,z) = D_{EC}(E_{NC}(T(x_i,E_{NC}(x_i,z)),z)).\\
\end{equation}
The input image $x_i$ and noise $z_i$ are encoded into a vectorial representation through $E_{NC}$. This representation is then passed to the spatial transformer network $T$, that generates an affine transformation and uses it to transform $x_i$. Finally, the transformed image is passed to a U-Net, composed of a convolutional encoder $E_{NC}$ and decoder $D_{EC}$.

%We first generate a representation $E_{NC}(x_i,z)$, which is given to the spatial transformer $T$ to generate an affine transformation of the original image $x_i$. This new image is then given as input to a U-Net~\cite{ronneberger2015u} to generate the final transformed image. %represented as a sequence of an encoder $E_{NC}$ and a decoder $D_{EC}$ network.
% We introduce the noise vector $z$ in the transformation, so that not only a single transformation per image is learned, but a distribution of transformations.

% Generator loss
The loss function of the generator can be formulated as the weighted sum of three terms:
\begin{equation}
\label{equ:gloss}
\begin{split}
\mathcal{L}_G = -\alpha\mathbb{E}_{x_i, y_i\sim p_{data},z \sim p_z} \left[ \log{(D^C(G(x_i, z), y_i))} \right]\\
- \beta\mathbb{E}_{x_i\sim p_{data,z \sim p_z}} \left[ \log{(D^D(x_i, G(x_i, z)))} \right]\\
- \gamma\mathbb{E}_{x_i, y_i\sim p_{data}} \left[ \log{(1-C_{y_i}(G(x_i,z))} \right],
\end{split}
\end{equation}
% where $G(x_i, z)$ is a transformation of the sample $x_i$ of label $y_i$ and a random noise vector $z$, 
in which $D^C$, $D^D$ are respectively the class and dissimilarity discriminator. Their role is to enforce constraints on the transformed image. More details will be given in the following paragraphs. %$D^C_{y_i}$ and 
$C_{y_i}$ is the softmax output of the classification network for the class $y_i$, \ie the probability that the given image belongs to the class $y_i$. %of the classifier for the class $y_i$.
Finally, $\alpha$, $\beta$ and $\gamma$ are hyper-parameters introduced to balance the three loss terms and stabilize the training of the model.

The first term of the loss function increases the probability $D^C$ that a transformed sample $G(x_i, z)$ does belongs to the same class $y_i$ as the original sample. The second term increases the probability $D^D$ that the transformed sample $G(x_i,z)$ and the original sample $x_i$ are different. Finally, the third term reduces the probability of a correct classification $C_{y_i}$ of the transformed sample $G(x_i,z)$.
%In contrast to previous GAN methods that use a classifier (\eg Triple GAN~\cite{chongxuan2017triple} and Bayesian DA~\cite{tran2017bayesian}), 
This loss pushes the generator to produce images that are difficult to classify and, if properly balanced, helps to improve the classifier (see Sec.~\ref{sub:ablation} Adversarial loss). %Previous GAN methods that use a classifier (\eg Triple GAN~\cite{chongxuan2017triple} and Bayesian DA~\cite{tran2017bayesian}) did not use this additional term.
%, the dissimilarity between the original image and the transformed image is maximized. Finally, in the third one, the probability of the real label for the transformed sample is minimized in order to make the classifier robust against adversarial samples. 
% Discriminators
% Class discriminator
\paragraph{Constraints on transformations.}
To enforce transformations that do not alter the class of an image, the generator is supported by two discriminators. The first one, the class discriminator $D^C$, ensures that the generated image belongs to the same class as the original image. The second, the dissimilarity discriminator $D^D$, ensures that the generated sample is different from the original sample.
% The motivation behind this design is that we want to create new samples that are as different as possible from the original samples but still in the same class distribution. 
%In other words, we want to increase the variance of the dataset.

The first discriminator, $D^C$, receives as input an image (either a real image $x_i$ or a transformed image $G(x_i,z)$) and a class label $y_i$ and outputs the probability of the image to belong to that class. Its loss function can be formulated as:
\begin{equation}
\begin{split}
\mathcal{L}_{D^C} = -\mathbb{E}_{x_i, y_i\sim p_{data}} \left[ \log{(D^C(x_i, y_i))} \right]\\
- \mathbb{E}_{x_i, y_i\sim p_{data},z \sim p_z} \left[ \log{(1 - D^C(G(x_i, z), y_i))} \right].
\end{split}
\end{equation}
% where $G(x_i, z)$ is a transformation of the sample $x_i$ of label $y_i$ and a random noise vector $z$.
The first term increases the probability $D^C(x_i, y_i)$ that a real sample $x_i$ belongs to class $y_i$, whereas the second term reduces the probability $D^C(G(x_i, z), y_i)$ that a generated sample $G(x_i,z)$ belongs to the same class $y_i$. In this way the discriminator learns to distinguish between real and generated samples of a certain class. 

%Dissimilarity discriminator
% \paragraph{Dissimilarity discriminator.}
 The second discriminator, $D^D$, takes a pair of samples as input (either two different samples of the same class $x_i,x_j$ or a sample $x_i$ and its transformation $G(x_i,z)$), and outputs a dissimilarity probability between the two samples. Its loss function can be formulated as:
\begin{equation}
\begin{split}
\mathcal{L}_{D^D} = -\mathbb{E}_{x_i,x_j \sim p_{data}} \left[ \log{(D^D(x_i, x_j))} \right]\\
-\mathbb{E}_{x_i\sim p_{data},z \sim p_z} \left[ \log{(1-D^D(x_i, G(x_i, z)))} \right]
\end{split}
\end{equation}
% where $G(x_i, z)$ is a transformation of the sample $x_i$ of label $y_i$ and a random noise vector $z$ and $x_j$ is a sample from the same class as $x_i$.
In the first term of the loss function increases the probability $D^D(x_i,x_j)$ of a sample $x_i$ and another different sample from the same class $x_j$, whereas the second term reduces the probability $D^D(x_i,G(x_i,z))$ between the same sample $x_i$ and the corresponding transformed sample $G(x_i,z)$. In this way this discriminator enforces that the transformed sample $G(x_i,z)$ belong to the same class as $x_i$ while being different (as we never use $x_j=x_i$). Thus, it enforces dissimilarity between the transformed sample and the original one.

% Classifier
\paragraph{Classification.}
The image classifier $C$ is trained jointly with the generator and the two discriminators. $C$ is fed with real samples $x_i$ as well as augmented samples, \ie samples transformed by $G$. Its loss function can be formulated as:
\begin{equation}
\begin{split}
\mathcal{L}_C = - \mathbb{E}_{x_i,y_i\sim p_{data}} \left[ \log{(C_{y_i}(x_i))} \right]\\
- \mathbb{E}_{x_i,y_i\sim p_{data},z \sim p_z} \left[ \log{(C_{y_i}(G(x_i,z)))} \right]
\end{split}
\end{equation}
%where $G(x_i, z)$ is a transformation of the sample $x_i$ of label $y_i$ and a random noise vector $z$.
%In the first term of the loss function, the cross entropy loss between the predicted labels of the true samples and the true label distribution is minimized whereas in the second term, the cross entropy loss between the predicted labels of the transformed samples and the true label distribution is minimized.
%same as above
This loss is a classical cross-entropy loss that enforces the classifier to give higher probability to the correct classification class in case of real $x_i$ or augmented samples $G(x_i,z)$
\paragraph{Global Loss.}
 Training our model consists in finding the equilibrium in a multiple two-player game. Indeed we solve jointly the adversarial game between $G$ and $D^C$, $G$ and $D^D$ and finally between $G$ and $C$. %In contrast to previous GAN methods that use a classifier (\eg Triple GAN~\cite{chongxuan2017triple} and Bayesian DA~\cite{tran2017bayesian}), we introduce an additional loss pushing the generator to produce images that are difficult to classify, and this also helps to improve the classifier.\\
% Finally, we want to minimize a global loss to find the optimal parameters for the generator, discriminators and classifier. 
The global loss can be formulated as:
\begin{equation}
\mathcal{L} = \mathcal{L}_G + \mathcal{L}_{D^C} + \mathcal{L}_{D^D} +\mathcal{L}_C
\end{equation}
During optimization, we sequentially minimize a mini-batch of each loss. % For additional details about the learning algorithm see the supplementary material.
Notice that $\mathcal{L}_G$ tries to maximize $D^C$ of the transformed samples $G(x_i,z)$, while $\mathcal{L}_{D^C}$ tries to maximize $1-D^C$, which corresponds to minimize $D^C$. The same also for $D^D$ and $C$. This is not a problem, in fact it shows that the defined loss is adversarial, in the sense that generator and discriminator/classifier ``fight'' to push the losses in different directions. %If the optimization is tuned properly, 
This mechanism generates augmented samples that help the training of the classifier, \ie samples that belongs to the right class but are close to the decision boundaries.

\paragraph{Similar Approaches.}
Our model has some similarities with Data Augmentation GAN~\cite{antoniou2018augmenting}, as we also generate transformations of images, instead of generating images from scratch. In our approach, however, we learn this transformation together with the classifier in a joint learning setting, allowing for improved classification performance, as show in Sec.~\ref{sub:joint}.

The joint training of a classifier and discriminator is not novel. If the aim is to produce a good classifier, having a separate classifier from the discriminator is important as pointed  in~\cite{dai2017good}. Similarly to our approach, Triple GAN~\cite{chongxuan2017triple} and Bayesian Data Augmentation GAN~\cite{tran2017bayesian} use a classifier separated from the discriminator and learned jointly with the generator. In our case however, instead of generating new images from scratch, we produce image transformations using two discriminators, one for the class, and a second one to avoid identity transformation. Additionally, we introduce a term in the loss (third term of $\mathcal{L}_G$ in equ.~\ref{equ:gloss}) to push the generator to produce images that are difficult to classify, and this also helps to improve the classifier.
Finally, we are the first to show that a spatial transformer network~\cite{jaderberg2015spatial} can be used in an adversarial way to improve the classification performance.

%%%%%%%%%%%%%%%%%%%%%%
% EXPERIMENTS
%%%%%%%%%%%%%%%%%%%%%%
\section{Experiments}
\label{sec:experiments}
We present several experiments to better understand our model and compare it with the state-of-the-art in automatic data augmentation.
In a first set of experiments, we compare the performance of learning data transformations to standard data augmentation for different sizes of the training datasets. In a second one, we compare the performance of our model to other state-of-the-art models. In a third one, we show the importance of learning jointly the data augmentation and the classifier, and finally, we analyze the contribution of each model component to the classifier accuracy.

\subsection{Datasets}
For our experiments, we tested our model on four well known datasets.
MNIST~\cite{lecun1998gradient} is a dataset of handwritten grayscale digits.
The full dataset contains 70,000 samples, 60,000 training samples and 10,000 test samples.
In our experiments with reduced dataset, we use a subset of 550 samples as in \cite{ratner2017learning}. 
SVHN~\cite{netzer2011reading} (Street View House Numbers) is a dataset of 32 x 32 pixels images of real world color photos of house numbers. It is composed of 73257 training samples and 26032 test samples.
In our experiments with reduced dataset, we use a subset of 1000 training samples.
CIFAR10~\cite{krizhevsky2009learning} is a dataset of 10 classes color natural images of size 32 x 32. 
The full dataset is composed of 60,000 images, 50,000 training images and 10,000 training images. 
In our experiments with reduced dataset, we use a subset of 4000 training samples as in \cite{ratner2017learning}.
Fashion-MNIST~\cite{xiao2017online} is a dataset of grayscale fashion articles. The full dataset is composed of 70,000 samples, 60,000 training samples and 10,000 test samples. Similarly to MNIST, we use a subset of 550 samples in our experiments with reduced dataset.
%We do not experiment on larger datasets as we are interested in the difficult case of training with a reduced amount of labelled data.  Also, 
We did not experiment on datasets with high resolution images due to the known difficulties of standard GAN models to generate good quality high resolution images. We leave experiments with more advanced GAN optimization and models as a future work.

\subsection{Implementation Details}
\label{subsec:implementation_details}
In all our experiments, we apply a basic pre-processing to the images, which consists in subtracting the mean pixel value, and then dividing by the pixel standard deviation.
%The generator is a combination of a STN~\cite{jaderberg2015spatial} module followed by a U-Net~\cite{ronneberger2015u} network.
The generator network takes as input an image and a Gaussian noise vector (100 dimensions), which are concatenated in the first layer of the network. 
The three parameters $\alpha,\beta$ and $\gamma$ of the generator loss are estimated on a validation set. 
For the class discriminator $D^C$, we use the same architecture as in \cite{dai2017good}. The network is adapted to take as input an image and a label (as a one hot vector). These are concatenated and given as input to the first layer of the architecture.
For the dissimilarity discriminator $D^D$, we also use the same architecture. The network is adapted to take as input a pair of images, which are concatenated in the first layer of the architecture. For the classifier, we use the architecture used in \cite{dai2017good}. We use Adam as optimizer.

\subsection{Comparison with Standard Data Augmentation}
In this series of experiments, we compare the data augmentation learned by our model to a standard pre-defined data augmentation.
In order to do this, we define two different levels of data augmentation. \emph{Light DA} refers to random padding of 4 pixels on each side of the image, followed by a crop back to the original image dimensions. \emph{Strong DA} includes the same augmentation as \emph{light DA} but also rotation in range [-10, 10] degrees, scaling, with factor in range [0.5, 2]. For CIFAR10, \emph{strong DA} also includes a horizontal image flip. 

In a first experiment we compare the accuracy of the baseline classifier without any data augmentation (\emph{Baseline}), the baseline with two levels of pre-defined data augmentation (\emph{Baseline + light DA} or \emph{Baseline + Strong DA}), and our data augmentation model (\emph{Our Model}) while increasing the number of training samples. For very few samples (1000) the predefined data augmentation is still better than our approach. When the generation of the samples is learned with a dataset too small, the generator produces poor samples that are not helpful for the classifier. When the number of samples increases, our approach obtains a much better accuracy than the approach with standard data augmentation. For instance, at 4000 training samples, the baseline obtains an accuracy of 66\%, the predefined data augmentation obtains an accuracy of 76\% and our model reaches an accuracy of 80.5\%, thus a net gain of 14 points compared with the baseline and 4 points compared to the data augmented model.
If we add more examples, the gap between our learned data augmentation and the standard data augmentation tends to reduce. With the full dataset we reach about a half a point better than the standard data augmentation.
\begin{figure}[t]
    \centering
\begin{tikzpicture}[scale=0.95]
  \begin{semilogxaxis}[ 
    legend cell align={left},
    width=1.1\linewidth,
    height=0.8\linewidth,
    xlabel=Training Samples,
    ylabel=Accuracy,
    grid=major,
    legend pos=south east,
    xlabel near ticks,
    xticklabel style={/pgf/number format/1000 sep=},
    ylabel near ticks,
    xtick=data,
    yticklabel style={
        left,
        /pgf/number format/.cd,
        fixed,
        precision=2,
        /tikz/.cd
    },
    enlarge y limits={value=.1,upper},
    log ticks with fixed point,
    ymin=40
  ] 
  \addplot[color=green,mark=o] coordinates { (1000,45.12) (2000, 55.85) (4000,66.23) (10000,76.89) (25000,84.25) (60000,89.87) };
  \addplot[color=orange,mark=o] coordinates { (1000,47.00) (2000,59.27) (4000,73.38) (10000,81.17) (25000,87.78) (60000,91.30) };  
  \addplot[color=blue,mark=o] coordinates { (1000,56.37) (2000,66.66) (4000,76.3) (10000,85.204) (25000,90.27) (60000,92.59) };
  \addplot[color=red,mark=o] coordinates { (1000,53.25) (2000,69.96) (4000,80.51) (10000,86.11) (25000,90.61) (60000,93.04) };
    \legend{Baseline, Baseline + light DA, Baseline + strong DA, Our Model}
  \end{semilogxaxis}
\end{tikzpicture}
    \caption{\textbf{Classification Accuracy vs number of training samples on CIFAR10.} Our method is effective when the number of samples is reduced. However for too few samples, normal data augmentation it is still slightly better.}
    \label{fig:my_label}
    \vspace{-5mm}
\end{figure}
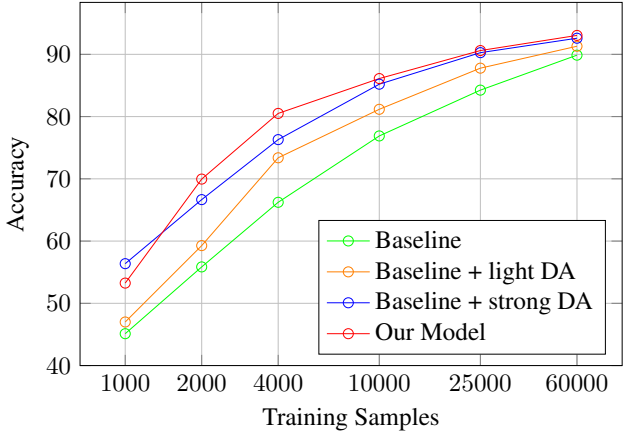
\begin{table}[H]
\centering
\resizebox{\columnwidth}{!}{%
\begin{tabular}{l|c|c|c|c}
     & \textbf{MNIST} & \textbf{FMNIST} & \textbf{SVHN} & \textbf{CIFAR10}\\
    \textbf{Method} & 550 & 550  & 1000 & 4000 \\
    \shline
    Baseline & 90.81 & 79.02  & 79.55 & 66.73\\
    Baseline + light DA & 97.55 & 78.96  & 84.48 & 74.76\\
    Baseline + strong DA & 98.50 & 80.37  & 84.33 & 77.74\\    
    Our model & 98.61 & 82.43  & 86.07 & 80.5\\
\end{tabular}%
}
\caption{\textbf{Comparison with DA on different datasets}. In low regime, our model performs better than \emph{light DA} and \emph{strong DA} on the four considered datasets. For more training data we expect a saturation of the improvement similar to Fig.~\ref{fig:my_label} for CIFAR10.}
\label{tab:comparison_DA_diffdatasets}
\vspace{-5mm}
\end{table}

In a second experiment, we compare different types of data augmentation on four datasets with a reduced number of samples.
As shown in Tab.~\ref{tab:comparison_DA_diffdatasets}, \emph{our best model} is always performing better than \emph{light DA} and \emph{strong DA}. This means that our data augmentation model learns transformations that are more useful for the final classifier. 
Notice that on FMNIST, \emph{light DA} decreases performance of the final classifier. This suggests that data augmentation is dataset dependent and transformations producing useful new samples in some domains might not be usable in others.

\subsection{Comparison with State of the Art}
\begin{table}[ht]
\centering
\resizebox{\columnwidth}{!}{%
\begin{tabular}{l|c|c|c|c}
& & \textbf{MNIST} & \textbf{CIFAR10}& \textbf{CIFAR10}\\
\textbf{Method} & \textbf{Model} & 550 & 4000 & Full\\
\shline
%VAT~\cite{miyato2018virtual} & ConvNet & - & 85.8$^*$ & 94.2 \\
%Ladder Network~\cite{rasmus2015semi} & ConvNet & $99.14^*$ & 79.91$^*$ & 90.73 \\ 
%CatGAN~\cite{springenberg2015unsup} & ConvNet  & 98.61$^*$ & 80.42$^*$ & 90.62\\ 
%\hline
%Improved GAN~\cite{} & ConvNet & - & 67.0 & - \\
Baseline & ConvNet & 90.81 & 66.23 & 89.88\\
Bayesian DA~\cite{tran2017bayesian} & ResNet18  & - & - & 91.0\\
DADA~\cite{zhang2018dada} & ResNet56 & - & 79.3 & - \\
TANDA~\cite{ratner2017learning}(MF) & ResNet56 & 96.5 & 79.5 & 94.4\\
TANDA~~\cite{ratner2017learning}(LSTM) & ResNet56 & 96.7 & 81.5 & 94.0\\
Our model & ConvNet & 96.0 & 80.50 & 93.0\\
\end{tabular}%
}
\caption{\textbf{Comparison to other automatic DA Methods.} We compare the accuracy of our model with other methods performing automatic data augmentation on MNIST and CIFAR10.
Notice that we use only labelled samples for the training, therefore the task is harder than in semi-supervised learning. }
\label{tab:comparison_SOTA}
\end{table}
In Tab.~\ref{tab:comparison_SOTA} we compare our method with other approaches for automatic data augmentation. Compared with \emph{TANDA}\cite{ratner2017learning}, our method obtains slightly lower accuracies. However, \emph{TANDA} is based on the selection of predefined transformations. This means that its learning is reduced to a set of manually selected transformation, which facilitates the task.  %, while our approach does not need to predefine any transformations. 
Also, \emph{TANDA} uses an additional standard data augmentation based on image crop, while our method does not need any additional data augmentation. %Finally, for the evaluation on 4,000 samples, TANDA uses all training data for better learning the composition of transformations, while in our model we only use 4,000 training samples.
On the other hand, our method compares favorably to \emph{Bayesian DA}~\cite{tran2017bayesian} and \emph{DADA}~\cite{zhang2018dada}, both based on GAN models with a larger neural network as classifier. This shows that our combination of global and local transformations helps to improve the final performance of the method.
Notice that our approach considers only fully supervised data, thus a direct comparison with semi-supervised methods, such as \cite{chongxuan2017triple,springenberg2016iclr,miyato2018virtual}, that make use of unlabelled data, would not be fair.

\subsection{Joint Training}
\label{sub:joint}
\pgfplotstableread[row sep=\\,col sep=&]{
    epochs & Baseline & Joint Training & Separate Training \\
    200     & 63.42 & 75.79 & 72.2 \\
    500     & 64.28 & 78.65 & 72\\
    700     & 65.09 & 78.68 & 71.1\\
    1000    & 66.23 & 80.50 & 64.5\\
    }\mydata

In this experiment, we compare the performance of our method, in case of joint and separate training.
In joint training the generator of augmented images and the classifier are trained simultaneously in an end-to-end training as explained in our method. In separate training instead, the generator is first trained to generate augmented images, and these images are then used as data augmentation to improve the classifier.
In case of separate training, we collect samples from different phases of the training: at epochs 200, 500, 700, 1000.
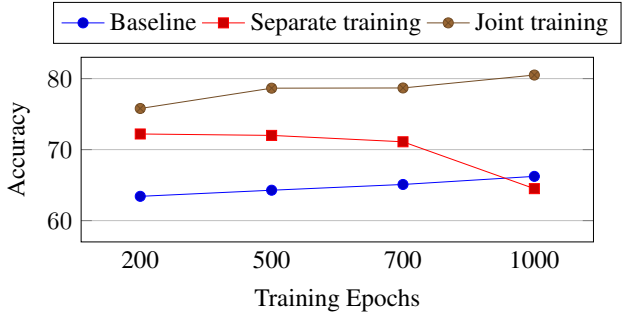
\begin{figure}
    \centering
\begin{tikzpicture}[scale=0.95]
\begin{axis}[ 
    width=1.05\linewidth,
    height=0.5\linewidth,
    ymajorgrids=true,
    xlabel=Training Epochs,
    ylabel=Accuracy, 
    legend style={at={(0.5,01.30)},
    anchor=north,legend columns=3},
    enlargelimits=0.15,
    symbolic x coords={200,500,700,1000},
    xtick=data,
    ylabel near ticks,
    ymax = 80,
    ymin = 60,
    xtick style={draw=none}, 
    xtick distance = 1000pt]
    \addplot table[color=green,mark=o,x=epochs,y=Baseline]{\mydata};
    \addplot table[color=blue,mark=o,x=epochs,y=Separate Training]{\mydata};
    \addplot table[color=red,mark=o,x=epochs,y=Joint Training]{\mydata};
    \legend{Baseline, Separate training , Joint training} 
\end{axis}
\end{tikzpicture}
\caption{\textbf{Classification Accuracy over epochs} on 4000 samples of CIFAR10 for a {baseline} classifier and our {joint training}. We compare them with a {separate} data augmentation training at 200, 500, 700 and 1000 training epochs.}
\label{fig:Joint}
\end{figure}
On Fig.~\ref{fig:Joint}, it is interesting to notice the different behaviour of the two methods. In the early phase of training, at epochs 200, both the \emph{Separate training} (beige line) and the \emph{Joint training} (red line) perform  above 70\%, whereas the \emph{Baseline} (blue line) is much lower. 
However, with additional training epochs, the performance of \emph{Separate training} decreases, while \emph{baseline} and \emph{joint training} accuracies increase.
From this experiment, it seems clear that for sample generation based on generic transformations (in contrast to predefined transformations as in \cite{ratner2017learning}), the joint training of the generator and the classifier is important for optimal performance. This may be why Data Augmentation GAN~\cite{antoniou2018augmenting} seems to work only with a very reduced set of examples. 

We believe that for good performance in data augmentation it is not just about generating plausible augmented samples, but also about generating the right samples at the right moment, as in curriculum learning~\cite{bengio2009curriculum}. Our understanding is that in the beginning of training, even poor generated samples can help to improve the optimization. However, towards the end of the training, only realistic samples are needed for improved accuracy. Notice also that if we use samples generated towards the end of training, the performance of the classifier drops even below the baseline model. This is probably due to the fact that the generated samples are too complex for the classifier and it cannot learn how to generalize on them. Also, consider that in case of separate training, it would not be possible to use the adversarial loss on the classifier $G_{ADV}$, which also helps to improve the final performance.

\subsection{Ablation Study}
\label{sub:ablation}
\begin{table}[!ht]
\centering
\resizebox{\columnwidth}{!}{%
\begin{tabular}{c|l|c}
    \textbf{Conf.} & \textbf{Components} & \textbf{Acc.} \\
    \shline
    (a) & $C$ & 66.73\\
    (b) & $C$ + $T$ & 64.75\\ %Non adversarial STN
    (c) & $C$ + $T$ + $D^C$ + $D^D$ & 70.62 \\
    (d) & $C$ + $D_{EC}$ + $D^C$ & 68.03 \\
    (e) & $C$ +  $E_{NC}$ + $D_{EC}$  + $D^C$ & 67.24 \\
    (f) & $C$ + $E_{NC}$ + $D_{EC}$ + $D^D$ & 67.11 \\
    (g) & $C$ + $E_{NC}$ + $D_{EC}$ + $D^C$ + $D^D$ &  73.82 \\
    (h) & $C$ + $E_{NC}$ + $D_{EC}$ + $T$ + $D^C$ + $D^D$ & 80.14\\
    (l) & $C$ + $E_{NC}$ + $D_{EC}$ + $T$ + $D^C$ + $D^D$ + $G_{ADV}$ & \textbf{80.51}\\
\end{tabular}}
\caption{\textbf{Ablation Study} on CIFAR-10 with 4000 training samples. 
%(a) CNN classifier; (b) standard distance transform; (c) adversarial distance transform; (d) 
$C$ = CNN classifier; $T$ = standard spatial transformer network applied to the image; $D_{EC}$ = CNN decoder, from noise or small code to image; $E_{NC} + D_{EC}$ = combination of CNN encoder and decoder to obtain a new image; $T + E_{NC} + D_{EC}$ = adversarial use of the STN; $D^C$ = Class Discriminator; $D^D$ = Dissimilarity Discriminator; $G_{ADV}$ = adversarial loss on the classifier. For more detail read the text.}
\label{tab:ablation}
\end{table}
In Tab.~\ref{tab:ablation} we evaluate the importance of the different components of our network on CIFAR10 with 4,000 training samples. First we evaluate the impact of each basic module used for data augmentation and then consider their combination. Notice how the classification accuracy goes from 66\% of a normal classifier \emph{(a)} to 80.5\% for our best model \emph{(l)}, without using any predefined data augmentation.

\paragraph{STN and U-Net.}
First, we compare the standard use of spatial transformer network $C+T$ \emph{(b)} as a transformation invariant approach with our adversarial approach $C+T+D^C+D^D$ \emph{(c)} to generate augmented samples.
In our approach the input image, together with a noise vector are encoded into a small representation that is passed to the spatial transformer network $T$, producing an affine matrix. The input image transformed with this affine matrix is then used as data augmentation for the classifier. While the standard spatial transformer network (STN) does not really help to improve results, our adversarial STN already improves the baseline accuracy by 4\%. The transformations generated by our adversarial STN are simply affine transformations that can help to improve the classifier. For more general transformations, able to independently change every pixel of an image we use a convolutional encoder-decoder model $E_{NC}+D_{EC}$ \emph{(g)} based on U-Net, thus improving model accuracy to  73.82\%. Finally, combining both transformations \emph{(h)} further improves the classification accuracy to 80.14\%.

\paragraph{Discriminators.}
We assess the contribution of the two discriminators, $D^C$ and $D^D$ used in our final model. As shown in Tab.~\ref{tab:ablation},  $D^C$ alone \emph{(e)} slightly improves  performance. This is probably because using only a class discriminator does not prevent the generator to generate the identity transformation, thus hindering the classifier. When using only the dissimilarity discriminator $D^D$ \emph{(f)}, the accuracy is also only slightly improved. The best performance is reached when the two discriminators work together \emph{(g)}, boosting the accuracy to 73.82\%.

\paragraph{Generation vs. Transformation.}
We compare our approach based on transforming an image with the direct generation of a new sample $D_{EC}$+$D^C$ \emph{(d)} as in Bayesian Data Augmentation~\cite{tran2017bayesian}. We see that this model is better than the basic classifier \emph{(a)} but still far from the performance levels obtained by our approach \emph{(g)}. This results makes us believe that transforming an image is simpler than generating an image from scratch. %That is possibly the reason that our approach works better than $D_{EC}$+$D^C$.

\paragraph{Adversarial loss.}
Finally, we evaluate the effect of adding a loss that enforces the generator to be adversarial to the classifier, \ie generate samples that are difficult for the classifier. This corresponds to the last term of $\mathcal{L}_G$ in equ.~\ref{equ:gloss}. As shown in Tab.~\ref{tab:ablation} \emph{(l)}  the contribution of this additional loss helps to further improve the final accuracy.

\subsection{Generated Transformations}
\begin{figure}[ht]
\centering
\begin{subfigure}[h]{.47\linewidth}
\centering
\includegraphics[width=\textwidth]{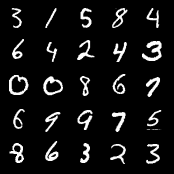}
\end{subfigure}
\hfill
\begin{subfigure}[h]{.47\linewidth}
\centering
\includegraphics[width=\textwidth]{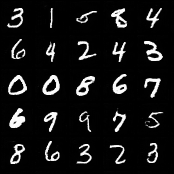}
\end{subfigure}
\vspace{0.2cm}
\centering
\begin{subfigure}[h]{.47\linewidth}
\centering
\includegraphics[width=\textwidth]{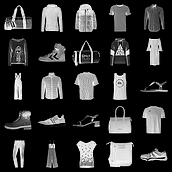}
\end{subfigure}
\hfill
\begin{subfigure}[h]{.47\linewidth}
\centering
\includegraphics[width=\textwidth]{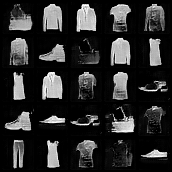}
\end{subfigure}
\vspace{0.2cm}
\centering
\begin{subfigure}[h]{.47\linewidth}
\centering
\includegraphics[width=\textwidth]{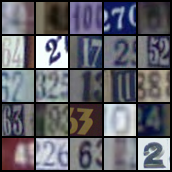}
\end{subfigure}
\hfill
\begin{subfigure}[h]{.47\linewidth}
\centering
\includegraphics[width=\textwidth]{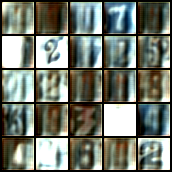}
\end{subfigure}
\vspace{0.2cm}
\centering
\begin{subfigure}[h]{.47\linewidth}
\centering
\includegraphics[width=\textwidth]{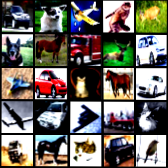}
\caption{Real samples}
\end{subfigure}
\hfill
\begin{subfigure}[h]{.47\linewidth}
\centering
\includegraphics[width=\textwidth]{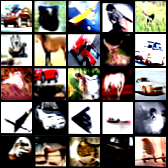}
\caption{Transformed samples}
\end{subfigure}
\caption{\textbf{Real and transformed images} from MNIST, Fashion-MNIST, SVHN and CIFAR10. Our approach learns to apply the right transformations for each dataset. For instance on MNIST and Fashion-MNIST there is no flip, nor zoom, because not useful, while on SVHN zoom is often used and on CIFAR10, both zoom, flip and color changes are applied. }
\label{fig:transform}
\end{figure}

In Fig.~\ref{fig:transform} we show some samples \emph{(left)} with the associated learned transformations \emph{(right)} for the four datasets considered. On MNIST, notice how the transformed numbers sometimes seem adversarial, in the sense that the applied transformation makes them look almost like another number. This is because hard samples are close to the boundaries between two or more classes and they are the most informative for improving the classification performance. For Fashion-MNIST it is interesting to notice that the transformation seems to reduce the variability of the dataset, transforming the objects into a set of simpler templates. Nevertheless, these transformations still help to improve the performance. On SVHN, only the digit in the center is used for classification. Interestingly, our model seems to learn that the generated samples are all zoomed to the central digit. Eventually, on CIFAR10 it is interesting to notice how the image colors and contrast are changed in a meaningful way. 

%%%%%%%%%%%%%%%%%%%%%%
% CONCLUSIONS
%%%%%%%%%%%%%%%%%%%%%%
\section{Conclusion}
\label{sec:conclusion}
In this work we have presented a new approach for improving the learning of a classifier through an automatic generation of augmented samples. The presented method learns general transformations end-to-end and is fully differentiable. In our experiments, we have shown that several elements are important to obtain the best performance.
First, the generator and the classifier should be trained jointly. Second, it is important to transform the given images instead of generating samples from scratch. Finally, the combined use of global transformations with STN and local transformation with U-Net is also essential to obtain the best data augmentation.

%%%%%%%%%%%%%%%%%%%%%%
% BIBLIOGRAPHY
%%%%%%%%%%%%%%%%%%%%%%
{\small
\bibliographystyle{ieee}
\bibliography{egbib}
}
\end{document}